\documentclass[11pt]{article}

\usepackage{algorithm}
\usepackage{algorithmic}
\usepackage[preprint]{acl}

\usepackage{times}
\usepackage{latexsym}

\usepackage[T1]{fontenc}

\usepackage[utf8]{inputenc}

\usepackage{microtype}

\usepackage{inconsolata}

\usepackage{graphicx}

\usepackage{booktabs}
\usepackage{multirow}

\usepackage{amsmath}

\usepackage{enumitem}
\setlist{nosep, leftmargin=*}

\usepackage{hyperref}

\title{Retrieval as Reasoning: Self-Evolving Agent-Native Retrieval via LLM-Wiki}

\author{
Haoliang Ming, Feifei Li, Xiaoqing Wu, Wenhui Que$^*$ \\
WeChat, Tencent Inc., Beijing, China \\
\texttt{\{hliangming, niyali, xiaoqingwwu, victorque\}@tencent.com}
}

\begin{document}
\maketitle

\begin{abstract}
LLM agents require retrieval to behave less like one-shot context fetching and more like reasoning: searching, reading, traversing, and deciding when evidence is sufficient. Yet current Retrieval-Augmented Generation (RAG) systems organize external knowledge as flat chunks retrieved by embedding similarity, exposing a retrieval-as-lookup interface ill-suited to iterative reasoning agents. We propose \textbf{LLM-Wiki}, an \textbf{agent-native retrieval system} that operationalizes the \textbf{Retrieval-as-Reasoning} paradigm by treating external knowledge as a compilable, composable, and self-evolving structure rather than a static retrieval index. LLM-Wiki compiles documents into structured Wiki pages with bidirectional links, exposes search, read, and link-following operations through standard tool-calling interfaces, and introduces an Error Book for persistent structural and semantic self-correction. LLM-Wiki achieves state-of-the-art results on HotpotQA, MuSiQue, and 2WikiMultiHopQA, outperforming HippoRAG 2, LightRAG, and GraphRAG by 2.0–8.1 F1 points. On AuthTrace, LLM-Wiki achieves the best overall accuracy, with especially strong gains on multi-document structured queries, confirming that compilation-based retrieval generalizes beyond chain-style multi-hop reasoning.
\end{abstract}

\section{Introduction}
\label{sec:intro}

Consider a 4-hop question from 2WikiMultiHopQA: ``Which film has the director who is older, \textit{The Gamecock} or \textit{Monster A Go-Go}?'' A dense retriever may retrieve the two film pages but miss the director biography pages containing birth dates, because those pages are semantically distant from the original query. Without this intermediate evidence, the system answers incorrectly. An agent with access to structured links, however, can read the film pages, follow explicit pointers to director biographies, and answer through compositional traversal. This example exposes a fundamental limitation of RAG \citep{lewis2020retrieval}: the bottleneck lies in how knowledge is organized and exposed to the agent, not merely in the retrieval algorithm. As LLM-based agents increasingly adopt ReAct-style \citep{yao2023react} tool-calling loops, they require knowledge that can be searched, read, and traversed as reasoning unfolds.

We call the prevailing pattern \textit{retrieval-as-lookup}: a retrieval module selects a fixed set of passages, a reasoning module consumes them, and retrieval itself does not adapt to the agent's intermediate observations. This motivates a shift to \textit{retrieval-as-reasoning}, in which an agent searches, inspects, follows intermediate entities, revises its retrieval plan, and stops only when the collected evidence is sufficient.

A central bottleneck is knowledge organization, not only retrieval.
Current RAG systems segment documents into flat chunks and store them in vector databases. This induces three limitations that better retrieval algorithms alone cannot address.

First, \textbf{flat chunks reduce retrieval to matching, not reasoning.} Selecting the top-$k$ fragments nearest a query embedding suffices for local fact lookup but grows brittle when an agent must follow relations, compare attributes, or aggregate cross-document evidence.

Second, \textbf{retrieval remains a single-shot black box, not an agent-controlled reasoning process.}
The system retrieves a fixed context before reasoning begins, so the agent cannot decide which entity to inspect next, follow newly discovered relations, or revise retrieval based on partial evidence.

Third, \textbf{structured knowledge bases degrade without self-correction.}
LLM-compiled knowledge bases can introduce dangling links, index inconsistencies, unsupported facts, and cross-page contradictions, yet existing pipelines lack a persistent loop for detecting, repairing, and reducing the recurrence of such errors.

Existing methods address parts of this picture but do not close the gap. RAPTOR and GraphRAG produce compressed summaries, HippoRAG~2 exposes KG triples, and LightRAG relies on entity and relation indices; these artifacts improve retrieval but do not provide a self-correcting, human-readable knowledge surface that an agent can plan over and traverse. Table~\ref{tab:method_comparison} in Appendix~\ref{app:method_comparison} gives a detailed comparison.

These limitations motivate three research questions: \textbf{(Q1)} Can compilation-based knowledge organization improve knowledge-intensive QA over flat RAG and graph-enhanced retrieval baselines?
\textbf{(Q2)} How does agent-controlled compositional traversal exploit Wiki structure through search, reading, link-following, and sufficiency checks?
\textbf{(Q3)} Can systematic structural and semantic compilation errors be detected, repaired, and mitigated through a persistent Error Book?

Guided by these questions, we propose LLM-Wiki, a system that operationalizes the Retrieval-as-Reasoning paradigm, in which the agent's retrieval process is itself a compositional reasoning activity: planning which knowledge units to visit, reading evidence, following compiled links, issuing revised searches when evidence is insufficient, and deciding when to answer. LLM-Wiki instantiates this paradigm through three mechanisms. First, it compiles raw documents into structured, interlinked Wiki pages rather than merely chunking and embedding them. Second, it exposes these pages through standard tool-calling interfaces so the agent can compose search, read, link-following, and sufficiency-checking operations. Third, it supports self-evolution through an Error Book that records systematic compilation errors, attributes root causes, converts them into reusable constraints, and drives both deterministic code fixes and periodic LLM-based repair.

LLM-Wiki is, to our knowledge, the first agent-native retrieval system that operationalizes the Retrieval-as-Reasoning paradigm. We summarize our contributions as follows:
\begin{enumerate}
    \item \textbf{The Retrieval-as-Reasoning paradigm via knowledge compilation and compositional traversal.}
    We propose LLM-Wiki, an agent-native retrieval system that compiles raw documents into structured, interlinked Wiki pages and transforms retrieval from fixed top-$k$ context selection into agent-planned compositional traversal.

    \item \textbf{Self-evolving Wiki knowledge through the Error Book.}
    We introduce a persistent self-correction mechanism that detects structural and semantic compilation errors, attributes root causes, accumulates reusable constraints, and drives two-layer repair across ingestion batches.

    \item \textbf{Empirical evidence that compilation-based retrieval scales with reasoning depth.}
    We evaluate LLM-Wiki on three public multi-hop QA benchmarks and AuthTrace, showing consistent gains over seven baselines, larger improvements as hop count increases, and the strongest overall performance on multi-document structured knowledge queries. Ablations further show that Wiki structure, progressive traversal, and the Error Book each contribute to the gains.
\end{enumerate}

\section{Related Work}
\label{sec:related}

\paragraph{Retrieval-Augmented Generation and Hierarchical Retrieval.}
RAG~\citep{lewis2020retrieval} augments language models with retrieved passages, while dense retrievers such as DPR~\citep{karpukhin2020dense} improve passage selection. For multi-hop reasoning, IRCoT~\citep{trivedi2023interleaving} and Self-RAG~\citep{asai2024selfrag} interleave retrieval with reasoning or self-reflection, while RAPTOR~\citep{sarthi2024raptor} and MemWalker~\citep{chen2023memwalker} organize documents into summary trees. However, these methods do not provide an explicitly linked structure over which an agent can plan compositional traversal paths.

\paragraph{Graph-Enhanced RAG and Knowledge Compilation.}
Graph-based systems such as GraphRAG~\citep{edge2024local}, HippoRAG~\citep{gutierrez2024hipporag}, HippoRAG~2~\citep{gutierrez2025hipporag2}, and LightRAG~\citep{guo2024lightrag} enrich retrieval with entities, relations, or graph-based indexing. Although they share our motivation of organizing knowledge beyond flat chunks, their outputs are typically summaries, triples, or vectorized graph structures rather than human-readable and agent-traversable pages. In contrast, LLM-Wiki compiles documents into Wiki pages with explicit bidirectional links. Table~\ref{tab:method_comparison} provides a detailed comparison.

\paragraph{LLM Self-Correction and Agent-Based Retrieval.}
LLM self-correction has been studied in Reflexion~\citep{shinn2023reflexion}, Self-Refine~\citep{madaan2023selfrefine}, and Self-RAG~\citep{asai2024selfrag}. These methods operate within a single inference episode or at query time, whereas the Error Book targets persistent cross-batch error patterns in knowledge construction through constraint accumulation and repair. More broadly, the LLM Wiki paradigm articulated by \citet{karpathy2026llmwiki} suggests compiling unstructured documents into persistent structured knowledge stores, but remains an abstract design pattern. We operationalize this paradigm through compilation, quality assurance, and compositional retrieval.

\section{Method}
\label{sec:method}

\begin{figure*}[t]
    \centering
    \includegraphics[width=\textwidth]{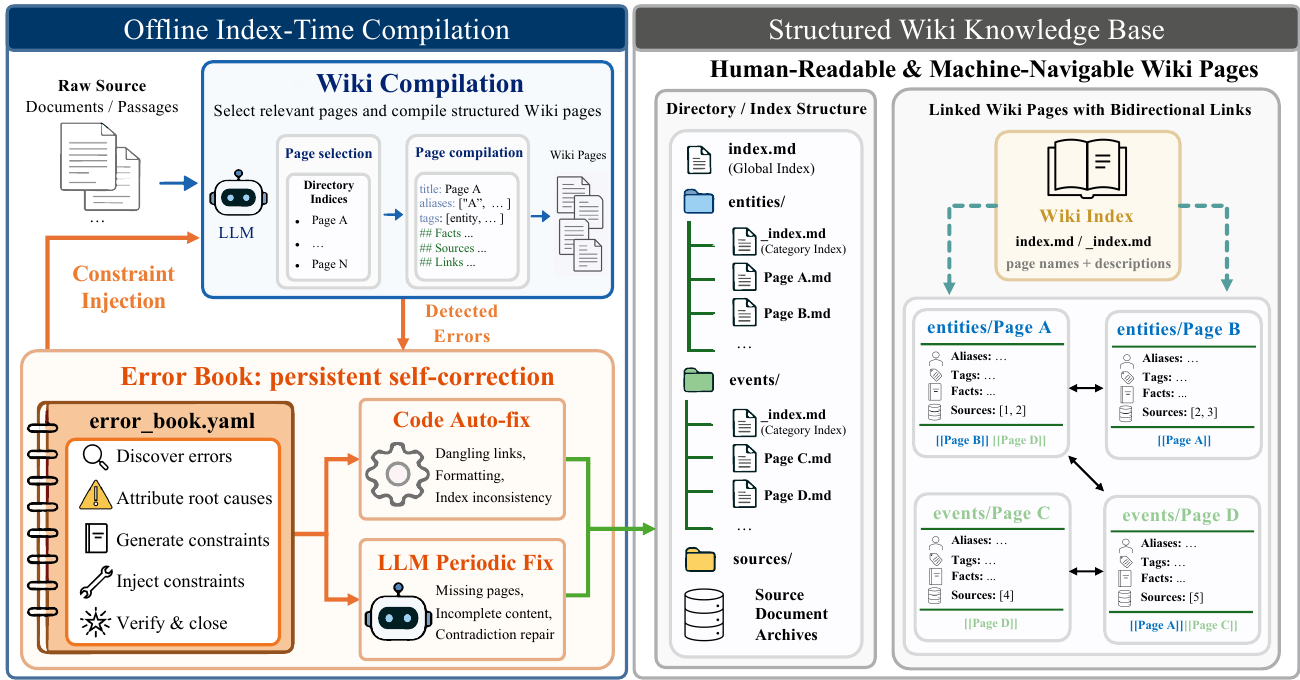}
    \caption{Index-time construction and structured Wiki knowledge base in LLM-Wiki. Raw documents are compiled into linked Wiki pages, while the Error Book records validation failures, injects constraints, and drives repair to produce a human-readable and machine-traversable Wiki.}
    \label{fig:overview}
\end{figure*}

Figure~\ref{fig:overview} illustrates LLM-Wiki's index-time construction. Raw documents are compiled into linked Wiki pages, while the Error Book records validation failures, injects constraints, and drives repair. The validated Wiki forms a human-readable and machine-traversable substrate for downstream traversal. Figure~\ref{fig:retrieval_compare} overviews the two traversal routes.

\paragraph{Retrieval-as-Reasoning Principles.}
We formalize the Retrieval-as-Reasoning paradigm through three principles that an agent-native retrieval system should satisfy:
(1) \textit{Compilability}---raw documents are transformed into structured, explicitly linked units that can be maintained as a persistent knowledge base;
(2) \textit{Composability}---retrieval is decomposed into atomic operations, such as search, read, and link following, that the agent composes through its reasoning loop; and
(3) \textit{Evolvability}---the knowledge structure self-corrects over time rather than silently degrading.
LLM-Wiki instantiates these principles through Wiki-structured knowledge, compositional retrieval, and the Error Book, respectively.
Together, these principles turn retrieval from a hidden ranking step into an explicit reasoning process over a maintained knowledge structure.

\subsection{Wiki-Structured Knowledge}
\label{sec:wiki_structure}

LLM-Wiki realizes \textit{compilability} by representing knowledge as a persistent Wiki rather than as independent chunks. The Wiki consists of directory indices, structured Markdown pages, and source archives. Each page exposes metadata, aliases, tags, facts, source references, and bidirectional wikilinks, allowing the agent to inspect overviews, locate pages, and follow links to related entities, events, concepts, or source digests.

At index time, LLM-Wiki compiles each source batch against the current Wiki state. For each passage, it selects relevant existing pages, generates structured Wiki updates, validates structural and content-level correctness, and passes detected errors to the Error Book. This incremental loop allows pages, links, and indices to evolve without rebuilding the entire knowledge base. Algorithm~\ref{alg:wiki_compilation} provides the full pseudo-code.

\subsection{Compositional Retrieval}
\label{sec:agent_loop}

LLM-Wiki realizes \textit{composability} through compositional Wiki traversal. Instead of receiving a fixed top-$k$ context, the agent composes \texttt{wiki\_search} and \texttt{wiki\_read} calls based on intermediate observations, iteratively searching, reading, following links, and checking sufficiency until it gathers sufficient evidence.

\begin{figure*}[t]
    \centering
    \includegraphics[width=\textwidth]{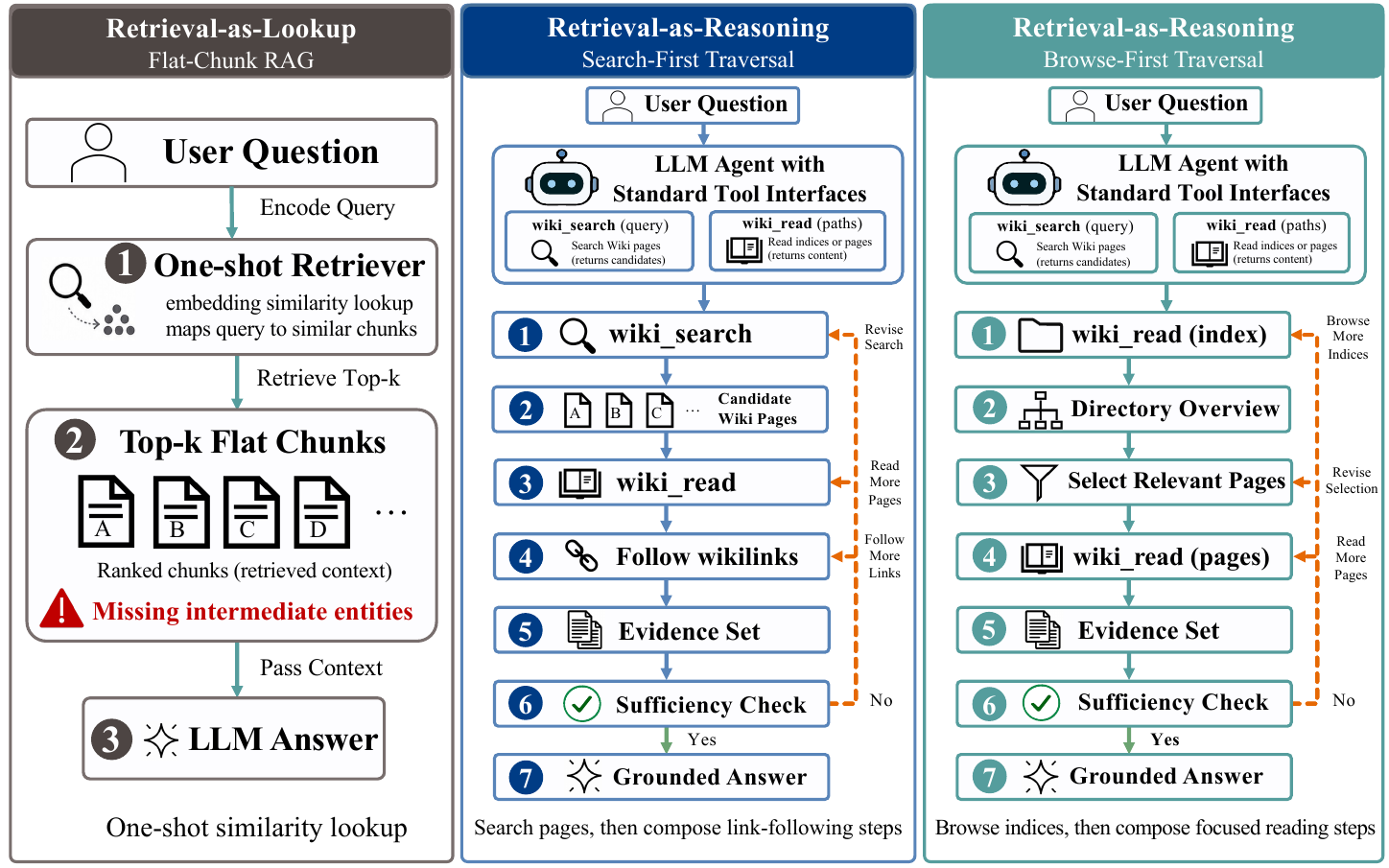}
    \caption{Retrieval-as-lookup versus retrieval-as-reasoning. Traditional flat-chunk RAG performs one-shot similarity lookup over independent chunks, whereas LLM-Wiki supports retrieval-as-reasoning through search-first and browse-first compositional traversal routes with evidence sufficiency checks and agent re-planning.}
    \label{fig:retrieval_compare}
\end{figure*}

Figure~\ref{fig:retrieval_compare} contrasts traditional flat-chunk RAG with two LLM-Wiki traversal routes: search-first traversal for entity-anchored queries and browse-first traversal for open-ended or enumeration queries.

\paragraph{Tool Interface.}
The agent is equipped with two tools:

\begin{itemize}
    \item \texttt{wiki\_search(query)}: Searches the Wiki index by prioritizing structured signals such as page names, aliases, tags, and descriptions before falling back to page content. It returns candidate pages and metadata for subsequent reading and traversal.
    \item \texttt{wiki\_read(paths)}: Batch-reads directory indices (\texttt{\_index.md}) or full pages. For knowledge pages, the returned content includes inter-page links that serve as traversal affordances for subsequent hops.
\end{itemize}

Search quality is less of a bottleneck than in single-shot RAG because the agent can judge page usefulness after reading and compensate for imperfect rankings through multi-round traversal.

\paragraph{Traversal Strategies.}
The agent adaptively selects retrieval strategies based on query characteristics:

\begin{itemize}
    \item \textbf{Direct access}: For known entities, the agent directly reads pages or first searches and then reads the top results.
    \item \textbf{Bridge queries} (A$\to$B$\to$answer): The agent reads page A, identifies entity B through inter-page links, and traverses to page B, reducing complex reasoning to iterative link traversal.
    \item \textbf{Exploratory browsing}: For open-ended queries, the agent reads directory indices to obtain a structured overview and then selectively reads promising pages.
\end{itemize}

\paragraph{Termination.}
After each \texttt{wiki\_read} call, the agent assesses evidence sufficiency. It terminates when all reasoning chains have been traced, the tool-call budget $T_{\max}$ is reached, or consecutive empty searches exceed patience threshold $P$. At least one \texttt{wiki\_read} call is required before answering.

\subsection{Error Book}
\label{sec:error_book}

LLM-Wiki realizes \textit{evolvability} through the Error Book, a persistent self-correction mechanism that detects recurring construction errors, attributes root causes, converts them into reusable constraints, and repairs affected pages. LLM-generated Wiki pages can contain structural and content-level issues, including dangling links, missing sections, malformed references, unsupported facts, factual inconsistencies, and cross-page contradictions. Traditional approaches rely on one-time post-processing or manual review, but the same error patterns may recur across ingestion batches. The Error Book improves the Wiki over time by injecting learned constraints into future compilation prompts and repairing existing errors.

\paragraph{Error Taxonomy.}
Through empirical analysis of compilation outputs across multiple corpora, we identify seven systematic error categories spanning structural validity, source grounding, and semantic consistency (Table~\ref{tab:error_types}). Structural errors include dangling links, incomplete pages, malformed references, unseen overwrites, and index inconsistencies, while content-level errors include unsupported facts and cross-page contradictions. Dangling links account for 29.1--63.8\% of detected errors across corpora, and malformed references account for 18.9--28.5\% (Table~\ref{tab:error_distribution}), indicating that link and source-reference validation are central to maintaining a traversable Wiki. Deterministic validators detect structural errors, whereas source-grounded LLM verification and cross-page consistency checks identify content-level errors. This distinction motivates our two-layer repair design: code-level fixes maintain the Wiki as a valid traversable structure, while LLM-based fixes improve factual grounding and cross-page consistency.

\paragraph{Self-Correction Loop.}
The Error Book operates through a five-stage lifecycle:

\begin{enumerate}
    \item \textbf{Discover}: After each compilation batch, deterministic validators detect structural errors, while source-grounded LLM verification and cross-page consistency checks identify content-level errors.
    \item \textbf{Attribute}: Each error is traced to its root cause, such as assuming a linked page exists without checking the index or over-generalizing from a source passage.
    \item \textbf{Constrain}: The root cause is formalized as a natural-language constraint rule, such as verifying link targets before emitting wikilinks or grounding entity attributes in cited source digests.
    \item \textbf{Inject}: All open constraint rules are appended to the Step~2 compilation prompt, guiding the LLM to avoid known error patterns in subsequent batches.
    \item \textbf{Verify \& Close}: Periodically, the system re-validates previously erroneous pages. If the error no longer appears after constraint injection, the error entry is marked as closed.
\end{enumerate}

The Error Book is persisted as a structured YAML file (\texttt{error\_book.yaml}), where each entry contains the error phenomenon, root cause analysis, generated constraint rule, verification method, and lifecycle status (\texttt{open}/\texttt{closed}). Constraints are injected as natural-language instructions, ranging from structural rules such as ``NEVER create a link to a page not present in \texttt{\_index.md}'' to semantic rules such as ``Do not add entity attributes unless they are supported by the cited source digest.'' These instructions leverage the LLM's instruction-following capability without requiring architectural modifications.

\paragraph{Two-Layer Repair: Code and LLM.}
Beyond constraint injection, the Error Book actively repairs existing errors through a two-layer mechanism. \textbf{Layer~1 (Code Auto-fix)} runs after every compilation batch and applies deterministic routines to fix structural errors, including dangling links, noisy formatting, and index inconsistencies. \textbf{Layer~2 (LLM Periodic Fix)} triggers every $N$ articles and handles semantic or content-level errors that require reasoning, such as missing pages, incomplete digests, unsupported facts, and cross-page contradictions. Thus, Layer~1 keeps the Wiki structurally valid, while Layer~2 improves factual grounding and cross-page consistency. At finalization, a three-round code-fix $\leftrightarrow$ LLM-fix loop catches newly introduced errors and drives the Wiki toward a stable state.

\section{Experimental Setup}
\label{sec:setup}

\subsection{Datasets}

We evaluate LLM-Wiki on three public multi-hop QA benchmarks and one structured-knowledge benchmark. For HotpotQA \citep{yang2018hotpotqa}, MuSiQue \citep{trivedi2022musique}, and 2WikiMultiHopQA \citep{ho2020constructing}, we use the first 500 QA examples in the original dataset order. The source corpus for each benchmark is the union of the context paragraphs associated with those 500 examples, and the same corpus is used for all methods. This is the dataset-provided context setting rather than open-domain full-Wikipedia retrieval.

We further evaluate on AuthTrace~\citep{wu2026authtracediagnosingevidenceconstruction}, a diagnostic benchmark for evidence selection and assembly in thematically dense single-author corpora. Each instance provides a query, quoted gold evidence units, atomic gold claim units, a reference answer, and an exact evidence fan-in label. Following the AuthTrace answer-correctness protocol, we use GPT-4o-mini to judge responses with a 0--3 rubric and report Single-doc, Low multi-doc, High multi-doc, and All settings.

\subsection{Baselines}

We compare against seven methods spanning four paradigms:

\begin{itemize}
    \item \textbf{None (Closed-book)}: Direct LLM generation without retrieval, serving as a lower bound.
    \item \textbf{Vanilla RAG (BM25)}: BM25 sparse retrieval + top-5 passages + LLM generation.
    \item \textbf{Vanilla RAG (Dense)}: Qwen3-Embedding-8B dense retrieval + top-5 passages + LLM generation.
    \item \textbf{RAPTOR} \citep{sarthi2024raptor}: Recursive clustering and summarization into a tree structure.
    \item \textbf{GraphRAG} \citep{edge2024local}: Entity extraction + community detection + hierarchical summarization.
    \item \textbf{LightRAG} \citep{guo2024lightrag}: Dual-level retrieval over entity and topic graphs.
    \item \textbf{HippoRAG~2} \citep{gutierrez2025hipporag2}: KG triples + Personalized PageRank + LLM QA.
\end{itemize}

\subsection{Metrics}

Public QA benchmarks use Answer F1 and Exact Match (EM), with hop-wise and type-wise F1 for fine-grained analysis. AuthTrace uses judged accuracy (AC, \%) under GPT-4o-mini with the benchmark's 0--3 answer-correctness rubric, normalized as $\mathrm{score}/3 \times 100$. Efficiency is measured by average query latency.

\subsection{Implementation Details}

All methods use GLM-5.1, from the GLM-5 model family~\citep{glm5}, as the unified answer LLM. For methods involving LLM-based indexing, summarization, graph construction, or Wiki compilation, we also use GLM-5.1 for those steps, so that differences primarily reflect retrieval and knowledge organization strategies rather than model choice. For all methods requiring text embeddings, including dense retrieval and graph-enhanced baselines, we use Qwen3-Embedding-8B~\citep{qwen3embedding}. For all baselines, we use the authors' official implementations and follow their recommended hyperparameters. All methods are evaluated on the same source corpus.

Our evaluation compares end-to-end retrieval systems rather than equalizing every retrieval component. LLM-Wiki uses a larger query-time tool-call budget, but we report latency in Table~\ref{tab:efficiency} and discuss its one-time compilation cost in the limitations.

For LLM-Wiki, the agent's maximum tool-call budget is $T_{\max}=15$, the patience threshold is $P=3$ consecutive empty searches, and \textsc{SelectPages} selects at most $k=5$ pages. The Error Book operates with a re-validation period of every 10 compilation batches.

\section{Experiments}
\label{sec:experiments}

\subsection{Main Results on Public Benchmarks}

\begin{table*}[t]
\centering
\setlength{\tabcolsep}{8pt}
\begin{tabular}{lcccccc}
\toprule
\multirow{2}{*}{\textbf{Method}} & \multicolumn{2}{c}{\textbf{HotpotQA}} & \multicolumn{2}{c}{\textbf{MuSiQue}} & \multicolumn{2}{c}{\textbf{2WikiMHQA}} \\
\cmidrule(lr){2-3} \cmidrule(lr){4-5} \cmidrule(lr){6-7}
& F1 & EM & F1 & EM & F1 & EM \\
\midrule
None (Closed-book) & 0.551 & 0.442 & 0.456 & 0.372 & 0.638 & 0.546 \\
Vanilla RAG (BM25) & 0.717 & 0.590 & 0.545 & 0.442 & 0.790 & 0.684 \\
Vanilla RAG (Dense) & 0.764 & 0.642 & 0.611 & 0.500 & 0.815 & 0.724 \\
RAPTOR & 0.801 & 0.674 & 0.522 & 0.442 & 0.707 & 0.652 \\
GraphRAG & 0.771 & 0.650 & 0.582 & 0.482 & 0.720 & 0.648 \\
LightRAG & 0.819 & 0.682 & 0.659 & 0.550 & 0.847 & 0.764 \\
HippoRAG~2 & 0.805 & 0.668 & 0.624 & 0.514 & 0.831 & 0.706 \\
\midrule
\textbf{LLM-Wiki (Ours)} & \textbf{0.839} & \textbf{0.710} & \textbf{0.739} & \textbf{0.634} & \textbf{0.911} & \textbf{0.854} \\
\bottomrule
\end{tabular}
\caption{Main results on three multi-hop QA benchmarks, with 500 questions per benchmark. Best results are shown in \textbf{bold}.}
\label{tab:main_results}
\end{table*}

As shown in Table~\ref{tab:main_results}, LLM-Wiki achieves the highest F1 and EM across all three datasets. The gains are largest on the more compositionally challenging benchmarks: on MuSiQue, LLM-Wiki outperforms Dense RAG by 12.9 F1 points and the strongest baseline, LightRAG, by 8.1 F1 points; on 2WikiMultiHopQA, it exceeds LightRAG by 6.4 F1 points. On HotpotQA, LLM-Wiki still achieves the best overall performance, with a 2.0 F1-point gain over LightRAG, indicating that it remains competitive even when strong single-shot retrieval baselines perform well. Overall, LLM-Wiki preserves performance on simpler 2-hop settings while its advantage increases with reasoning depth and compositionality.

Among the baselines, LightRAG performs strongest overall, whereas RAPTOR and GraphRAG consistently underperform. Both rely on hierarchical summarization, which can discard precise entity names and relations required for compositional reasoning: RAPTOR's recursive summaries may lose specificity, while GraphRAG's community detection groups entities by co-occurrence rather than explicit semantic relations. These results suggest that the key advantage of LLM-Wiki is not simply retrieving more text, but exposing knowledge in a form that lets the agent turn intermediate observations into follow-up retrieval actions.

\subsection{AuthTrace Results}

Table~\ref{tab:authtrace} reports judged accuracy (AC) on AuthTrace. LLM-Wiki achieves the best overall performance, outperforming the strongest baseline, HippoRAG~2, by 2.1 AC points. Its advantage increases as queries require broader cross-document reasoning: LLM-Wiki exceeds HippoRAG~2 by 5.0 AC points on Low multi-doc questions and by 8.9 AC points on High multi-doc questions. These results indicate that compilation-based knowledge organization is especially effective for gathering, comparing, and aggregating evidence across dispersed documents.

On Single-doc questions, HippoRAG~2 outperforms LLM-Wiki by 2.3 AC points. This is expected because many single-document questions can be answered once the relevant original article is retrieved and a local detail is identified. In contrast, LLM-Wiki answers from structured pages that reorganize source content into concise facts and summaries, which benefits cross-document synthesis but can occasionally omit fine-grained local details. Overall, AuthTrace shows that LLM-Wiki extends beyond standard multi-hop QA, with the strongest gains on structured queries requiring multi-document synthesis.

To validate the reliability of GPT-4o-mini-based judging, we additionally conduct a human audit on a random subset of AuthTrace outputs. Human judgments show high agreement with GPT-4o-mini and preserve the same ranking among audited methods; the protocol and results are reported in Table~\ref{tab:human_audit}.

\begin{table}[t]
\centering
\small
\setlength{\tabcolsep}{3pt}
\begin{tabular*}{\columnwidth}{@{\extracolsep{\fill}}lcccc}
\toprule
\textbf{Method} & \textbf{Single} & \textbf{Low} & \textbf{High} & \textbf{All} \\
\midrule
None (Closed-book)  & 12.3 & 16.8 & 16.2 & 14.0 \\
Vanilla RAG (BM25)  & 75.4 & 48.5 & 30.4 & 62.7 \\
Vanilla RAG (Dense) & 72.9 & 49.0 & 35.0 & 61.9 \\
RAPTOR              & 69.5 & 44.0 & 33.9 & 58.6 \\
GraphRAG            & 55.8 & 35.1 & 26.7 & 46.8 \\
LightRAG            & 73.4 & 34.2 & 14.5 & 55.7 \\
HippoRAG~2          & \textbf{78.3} & 59.6 & 46.5 & 68.3 \\
\midrule
\textbf{LLM-Wiki (Ours)} & 76.0 & \textbf{64.6} & \textbf{55.4} & \textbf{70.4} \\
\bottomrule
\end{tabular*}
\caption{Results on AuthTrace across fan-in settings. Scores are judged accuracy (AC). Best results in each column are shown in \textbf{bold}.}
\label{tab:authtrace}
\end{table}

\subsection{Ablation Study}

To isolate the contribution of each component, we conduct ablation experiments by removing one component at a time, as shown in Table~\ref{tab:ablation}.

\begin{table}[t]
\centering
\small
\setlength{\tabcolsep}{3pt}
\begin{tabular}{lccc}
\toprule
\textbf{Variant} & \textbf{HotpotQA} & \textbf{MuSiQue} & \textbf{2Wiki} \\
\midrule
\textbf{LLM-Wiki (full)} & \textbf{0.839} & \textbf{0.739} & \textbf{0.911} \\
\midrule
w/o Wiki Structure & 0.778 & 0.669 & 0.844 \\
w/o Progressive Traversal & 0.722 & 0.601 & 0.789 \\
w/o Error Book & 0.801 & 0.699 & 0.877 \\
\bottomrule
\end{tabular}
\caption{Ablation study using F1 scores. Each row removes one component while keeping the others intact.}
\label{tab:ablation}
\end{table}

\noindent\textbf{Removing Wiki structure.}
This variant removes the compiled Wiki representation and uses flat source chunks as the retrieval substrate while retaining the same agent framework and answer generation model. The agent can still issue retrieval calls, but the returned evidence consists of independent chunks rather than directory indices, structured pages, and wikilinks. The consistent drop of 6.1--7.0 F1 points indicates that structured knowledge organization provides the compiled substrate needed for effective retrieval and reasoning.

\noindent\textbf{Removing progressive traversal.}
This variant keeps the compiled Wiki unchanged but disables iterative re-planning. The agent performs a single \texttt{wiki\_search} call, reads the top retrieved pages once, and answers without following additional wikilinks or revisiting search/read steps. The 11.7--13.8 F1-point drop shows that the compiled Wiki structure is most useful when actively traversed: indices, pages, and links help because the agent can search, read, and follow them across steps, not because multi-round retrieval alone is sufficient.

\noindent\textbf{Removing the Error Book.}
This variant uses the same Wiki compilation and traversal pipeline but disables Error Book updates, constraint injection, and repair. Compilation errors remain observable during validation, but they are not accumulated into reusable constraints or used to repair future batches. The 3.4--4.0 F1-point drop suggests that the Error Book improves downstream retrieval by reducing the impact of recurring structural and semantic construction errors.

Overall, these ablations show that LLM-Wiki's gains arise from mutually reinforcing components. Wiki structure provides the compiled substrate, progressive traversal exploits this substrate for compositional evidence gathering, and the Error Book maintains the quality of the structure being traversed.

\subsection{Fine-Grained Analysis}

Fine-grained results further show that LLM-Wiki's gains are concentrated on queries requiring deeper relational reasoning. On 2WikiMHQA, the F1 gap over LightRAG widens from 5.7 F1 points on 2-hop questions to 8.3 F1 points on 4-hop questions, and LLM-Wiki reaches 0.983 F1 on 4-hop questions (Figure~\ref{fig:hopwise_app}). Type-wise results show especially strong gains on compositional questions, where LLM-Wiki outperforms Dense RAG by 15.6 F1 points and LightRAG by 7.3 F1 points (Table~\ref{tab:typewise_app}). Under GPT-4o, LLM-Wiki also maintains gains of 5.1--16.9 F1 points over Dense RAG (Table~\ref{tab:generalization}), suggesting that the benefit comes from knowledge organization rather than a specific answer model. Table~\ref{tab:efficiency} shows that these gains do not require slower query-time inference: LLM-Wiki is comparable to BM25/Dense RAG and substantially faster than LightRAG, HippoRAG~2, and RAPTOR on most benchmarks. Detailed analyses and retrieval traces are provided in Appendices~\ref{app:additional_analysis} and~\ref{app:case_study}.

Overall, these results provide empirical support for the Retrieval-as-Reasoning hypothesis: when retrieval is organized as agent-driven compositional traversal through compiled knowledge structures, performance scales with the reasoning depth required by the query.

\section{Discussion}
\label{sec:discussion}

\paragraph{When Does Wiki Beat RAG?}
LLM-Wiki is most valuable when queries require relation following, evidence enumeration, cross-document aggregation, or dynamic knowledge exploration. Its advantage increases with hop count and compositionality because pre-compiled links make intermediate entities explicitly discoverable. In contrast, source-localized questions may favor methods that retrieve the original article directly, which helps explain why HippoRAG~2 performs better on AuthTrace Single-doc questions.

\paragraph{Retrieval as Reasoning through Compilation.}
More broadly, our results clarify why the retrieval-as-reasoning distinction matters. The central improvement is not a stronger similarity function, but a different contract between the knowledge base and the agent: compilation makes the search space explicit, compositional tools let observations drive subsequent retrieval steps, and the Error Book keeps the structure reliable. Front-loading organization to compile time replaces runtime matching with a richer agent–knowledge interaction surface while also reducing query latency.

\section{Conclusion}
\label{sec:conclusion}

We presented LLM-Wiki, the first agent-native retrieval system to realize Retrieval-as-Reasoning. The results support three claims. First, compiling documents into linked Wiki pages improves knowledge-intensive QA over flat-chunk and graph-enhanced retrieval, with gains increasing as reasoning depth grows. Second, agent-planned compositional traversal yields additive gains beyond the compiled structure alone. Third, the Error Book mitigates recurring structural and semantic compilation errors and improves downstream robustness.

Beyond this specific setting, our results establish a design principle for agent-native retrieval: compile knowledge into navigable structure, expose it for compositional traversal, and maintain it through closed-loop self-correction—a paradigm we believe extends naturally to knowledge-intensive agent tasks beyond QA.

\section*{Limitations}

Two primary limitations warrant discussion. First, \textbf{compilation cost}: each source passage requires \textsc{SelectPages} and \textsc{CompileWikiPages}, making initial construction more expensive than chunk-and-embed approaches, though the cost is amortized over later queries. Second, \textbf{scalability}: as the Wiki grows to tens of thousands of pages, directory indices may become unwieldy and page selection may degrade. Web-scale and frequently changing corpora will require hierarchical indexing, sharding, stale-fact handling, and global directory maintenance. Extending LLM-Wiki to large-scale dynamic maintenance, multi-modal Wikis, and cross-corpus transfer remains future work.

\bibliography{llm_wiki_refs}

% ============================================================
% Appendix
% ============================================================
\appendix

\section{Additional Related Work and Paradigm Comparison}
\label{app:method_comparison}

This appendix provides additional details on how LLM-Wiki differs from prior retrieval and knowledge organization paradigms. Table~\ref{tab:method_comparison} summarizes representative methods in terms of their index product, knowledge form, and core bottleneck.

We compare methods along three axes. The index product describes the artifact produced after offline processing, such as chunks, summaries, triples, vectors, or Wiki pages. The knowledge form characterizes whether the artifact is primarily a compressed text representation, a machine-readable retrieval structure, or a human-auditable knowledge organization. The core bottleneck highlights the main limitation for agent-driven compositional retrieval. This comparison emphasizes that the key distinction is not whether a method uses a graph or an LLM, but whether the resulting knowledge base exposes explicit, traversable structure that an agent can search, read, and traverse.

\begin{table*}[h]
\centering
\resizebox{\textwidth}{!}{%
\begin{tabular}{llll}
\toprule
\textbf{Method} & \textbf{Index Product} & \textbf{Knowledge Form} & \textbf{Core Bottleneck} \\
\midrule
Vanilla RAG & flat chunks & embedding-indexed passages & cannot traverse intermediate nodes \\
RAPTOR / MemWalker & summary tree & text compression & tree lacks lateral associations \\
GraphRAG & community summaries & hierarchical compression & summaries lose detail; high cost \\
HippoRAG~2 & KG triples & machine-readable fragments & lossy triples; PPR is approximate \\
LightRAG & entity/relation vectors & vector index & cannot discover intermediate entities \\
\midrule
\textbf{LLM-Wiki (Ours)} & \textbf{Wiki pages + links} & \textbf{structured knowledge} & \textbf{compilation cost (one-time)} \\
\bottomrule
\end{tabular}%
}
\caption{Comparison of knowledge organization paradigms. Unlike existing methods that produce compressed summaries or machine-readable fragments, LLM-Wiki compiles documents into structured, human-auditable Wiki pages with explicit bidirectional links.}
\label{tab:method_comparison}
\end{table*}

\section{Dataset Details}
\label{app:datasets}

\paragraph{Public Multi-Hop QA Benchmarks.}
We evaluate on three established public benchmarks. HotpotQA \citep{yang2018hotpotqa} contains 2-hop questions requiring bridge reasoning or comparison over Wikipedia paragraphs. MuSiQue \citep{trivedi2022musique} contains 2--4-hop compositional questions with an anti-shortcut design that prevents single-hop answering. 2WikiMultiHopQA \citep{ho2020constructing} contains 2-hop and 4-hop questions spanning bridge, comparison, compositional, and inference types. For each dataset, we take the first 500 QA examples in the original dataset order and build the retrieval corpus from the union of their associated context paragraphs. This corpus is shared by all methods; for example, the 2WikiMultiHopQA evaluation corpus contains 3,440 passages. We do not retrieve from the full Wikipedia dump in these experiments.

\paragraph{AuthTrace.}
Existing public multi-hop benchmarks often emphasize heterogeneous evidence, bridge entities, and short-span answers. To test evidence construction under high topical similarity, we evaluate on AuthTrace~\citep{wu2026authtracediagnosingevidenceconstruction}, a benchmark built from 2,099 filtered QA instances over 860 public-domain writings by five modern Chinese prose writers. Each instance contains a query with title leakage removed, quoted gold evidence units, atomic gold claim units, a concise reference answer, and an exact evidence fan-in label. Fan-in groups instances into Single-doc (=1), Low multi-doc (2--3), and High multi-doc ($\geq$4), testing local grounding, small-scale aggregation, and broad synthesis. Following the AuthTrace answer-correctness protocol, we use GPT-4o-mini as the judge. The judge receives the query, gold claim units, a reference answer, and the predicted answer; it labels each gold claim as supported, partial, missing, or contradicted, checks irrelevant or erroneous extra content, and assigns a 0--3 score, which is normalized as $\mathrm{score}/3 \times 100$ for AC. The reference answer is used only to interpret the claim units.

\section{Human Audit of AuthTrace Judging}
\label{app:human_audit}

This appendix reports a human audit of the GPT-4o-mini-based AuthTrace judge used in the main experiments. The goal is to verify that automatic judgments are aligned with human assessment and preserve the same relative method ranking.

\paragraph{Audit Protocol.}
We sample 240 responses, stratified to cover 3 representative methods (LLM-Wiki, HippoRAG~2, and Dense RAG) and the 3 AuthTrace fan-in settings (Single-doc, Low multi-doc, High multi-doc). Three authors independently annotate each response using the same 0--3 answer-correctness rubric as the GPT-4o-mini judge. Annotators judge the coverage of gold claim units, penalize irrelevant or erroneous extra content, and assign a 0--3 score, where 3 indicates a fully correct answer and 0 indicates an incorrect answer. Annotators are blind to both the method identity and the GPT-4o-mini judgment. Disagreements are resolved by discussion to obtain a single human score per response.

\paragraph{Results.}
Table~\ref{tab:human_audit} reports human--GPT agreement and human-derived accuracy under the same fan-in grouping as Table~\ref{tab:authtrace}. Human and GPT-4o-mini judgments agree on 89.6\% of audited responses, with Cohen's $\kappa=0.79$, indicating substantial agreement~\citep{landis1977kappa}.

\begin{table}[h]
\centering
\small
\setlength{\tabcolsep}{4pt}
\begin{tabular}{lcccc}
\toprule
\textbf{Metric} & \textbf{Single} & \textbf{Low} & \textbf{High} & \textbf{All} \\
\midrule
Human--GPT agreement (\%) & 91.2 & 89.4 & 87.9 & 89.6 \\
Cohen's $\kappa$ & 0.82 & 0.79 & 0.76 & 0.79 \\
\midrule
LLM-Wiki human AC & 74.1 & \textbf{62.4} & \textbf{53.6} & \textbf{68.3} \\
HippoRAG~2 human AC & \textbf{76.5} & 57.9 & 44.7 & 66.4 \\
Dense RAG human AC & 70.8 & 47.2 & 33.4 & 60.0 \\
\bottomrule
\end{tabular}
\caption{Human audit of GPT-4o-mini-based AuthTrace judging on 240 stratified samples. Human--GPT agreement and Cohen's $\kappa$ indicate substantial agreement, and the human-derived ranking among audited methods matches the GPT-4o-mini ranking in Table~\ref{tab:authtrace} on every column.}
\label{tab:human_audit}
\end{table}

\paragraph{Discussion.}
Two observations follow. First, compared with the GPT-4o-mini AC scores in Table~\ref{tab:authtrace}, human-derived AC scores for the audited methods are 1.6--2.2 points lower, indicating that GPT-4o-mini is slightly more permissive under our rubric. This is consistent with prior work showing strong but imperfect alignment between LLM judges and human evaluation~\citep{zheng2023judging}. Second, the per-column ranking among audited methods is preserved: LLM-Wiki remains strongest on Low multi-doc, High multi-doc, and All, while HippoRAG~2 retains its Single-doc advantage. Agreement also remains high across fan-in settings (87.9--91.2\%). We therefore use GPT-4o-mini as the primary AuthTrace judge and treat the human audit as validation of the judging protocol.

\section{Compilation Algorithm}
\label{app:algorithm}

Algorithm~\ref{alg:wiki_compilation} gives the full index-time compilation loop used by LLM-Wiki, including page selection, Wiki update generation, validation, Error Book updates, and two-layer repair.

\begin{algorithm}[h]
\small
\caption{Index-time Wiki compilation}
\label{alg:wiki_compilation}
\begin{algorithmic}[1]
\REQUIRE Source batch $X$, current Wiki $W$, directory indices $I$, source archives $A$, active Error Book constraints $C$
\ENSURE Updated Wiki $W$ and Error Book $\mathcal{B}$
\FOR{each source passage $x \in X$}
    \STATE $S \leftarrow \textsc{SelectPages}(x, I)$ \COMMENT{select relevant existing pages}
    \STATE $U \leftarrow \textsc{CompileWikiPages}(x, S, C)$ \COMMENT{generate pages, links, and index updates}
    \STATE $E_s \leftarrow \textsc{StructuralValidate}(U, W)$
    \STATE $E_c \leftarrow \textsc{ContentValidate}(U, W, A)$
    \STATE $E \leftarrow E_s \cup E_c$
    \IF{$E \neq \emptyset$}
        \STATE $\mathcal{B} \leftarrow \textsc{UpdateErrorBook}(\mathcal{B}, E)$
        \STATE $C \leftarrow \textsc{ActiveConstraints}(\mathcal{B})$
        \STATE $U \leftarrow \textsc{CodeAutoFix}(U, E_s)$
    \ENDIF
    \STATE $W \leftarrow \textsc{ApplyUpdates}(W, U)$
\ENDFOR
\IF{\textsc{PeriodicFixDue}$(\mathcal{B})$}
    \STATE $W \leftarrow \textsc{LLMPeriodicFix}(W, \mathcal{B})$
    \STATE $\mathcal{B} \leftarrow \textsc{VerifyAndClose}(\mathcal{B}, W)$
\ENDIF
\RETURN $W, \mathcal{B}$
\end{algorithmic}
\end{algorithm}

\section{Example Wiki Structure and Page Schema}
\label{app:schema}

This appendix illustrates the directory organization and page format of a compiled LLM-Wiki instance. The examples are adapted from the Wiki generated for the 2WikiMHQA evaluation corpus, with long lists shortened for readability.

\paragraph{Directory Layout.}
A compiled Wiki is a self-contained Markdown file tree. The root contains a global \texttt{index.md} that provides a knowledge overview and directory catalog. Each knowledge directory contains a \texttt{\_index.md} file, which serves as a browsable page listing, and individual knowledge pages. A parallel \texttt{sources/} tree stores paragraph-level digests and full original articles for provenance. The 2WikiMHQA Wiki contains 5,825 knowledge pages across 6 thematic directories, plus 6,840 source pages:

\begin{scriptsize}
\begin{verbatim}
wiki/
  index.md                          
  concepts/                         
    _index.md
    Dynastic-Succession.md
    ...
  geography/                        
    _index.md
    ...
  history/                          
    _index.md
    ...
  media/                           
    _index.md
    ...
  organizations/                    
    _index.md
    ...
  people/                           
    _index.md
    John-V-Prince-of-Anhalt-Zerbst.md
    Ernest-I-Prince-of-Anhalt-Dessau.md
    ...
  sources/                          
    digests/                        
      john-v-prince-anhalt-zerbst.md
      ...
    articles/                       
      john-v-prince-anhalt-zerbst.md
      ...
\end{verbatim}
\end{scriptsize}

\paragraph{Global Index (\texttt{index.md}).}
The global index begins with a knowledge overview paragraph summarizing the Wiki's scope and domain coverage, followed by a directory catalog:

\begin{scriptsize}
\begin{verbatim}
# Wiki Directory Overview

> Knowledge Overview (updated xxxx-xx-xx)
> This high-density, relational encyclopedic
  repository functions as a sophisticated engine
  for complex multi-hop question answering ...

## Directory Catalog
- concepts/ -- theories, methods,
    genres, categories, abstract ideas
- geography/ -- cities, villages,
    countries, airports, landmarks
- history/ -- historical events,
    periods, battles, medieval history
- media/ -- films, albums, songs,
    TV shows, books, creative works
- organizations/ -- schools,
    universities, bands, companies, dynasties
- people/ -- biographies of
    historical figures, artists, politicians
- sources/ -- paragraph digests
    and original archives
\end{verbatim}
\end{scriptsize}

\paragraph{Directory Index (\texttt{\_index.md}).}
Each directory's \texttt{\_index.md} organizes pages into semantic sections with aliases, tags, and one-line summaries. For example, \texttt{people/\_index.md} groups its 3,241 entries under headings such as ``German Nobility,'' ``Chinese Film Directors,'' ``Brazilian Politicians,'' and ``Spanish Politicians,'' enabling efficient browsing:

\begin{scriptsize}
\begin{verbatim}
# People
> xx pages

## German Nobility
- [[Ernest-I-Prince-of-Anhalt-Dessau]]
  (Ernest I of Anhalt-Dessau, Ernst I von
  Anhalt-Dessau) #German-nobility
  #House-of-Ascania #Prince
## Chinese Film Directors
- [[Zhang-Yimou]] (Zhang Yimou)
  -- Prominent Chinese film director ...
  #director #chinese-cinema #fifth-generation
## Brazilian Politicians
- [[Adalberto-Pereira-dos-Santos]]
  (Adalberto Pereira dos Santos, General
  Adalberto) #Brazil #general #politician
\end{verbatim}
\end{scriptsize}

\paragraph{Knowledge Page.}
Each knowledge page follows a fixed schema: YAML frontmatter with type, creation and update timestamps, aliases, and tags; a one-line blockquote summary; structured key facts; bidirectional wikilinks to related pages; and source references. Below is the complete page for John~V, Prince of Anhalt-Zerbst:

\begin{scriptsize}
\begin{verbatim}
---
type: people
created: xxxx-xx-xx
updated: xxxx-xx-xx
aliases: [John V of Anhalt-Zerbst,
  Johann V von Anhalt-Zerbst,
  Prince John V of Anhalt-Zerbst]
tags: [German nobility, House of Ascania,
  Prince, Anhalt-Zerbst, Anhalt-Dessau]
---
# John V, Prince of Anhalt-Zerbst
> German prince of the House of Ascania who
> ruled Anhalt-Dessau and later the re-created
> principality of Anhalt-Zerbst from 1544

## Key Facts
- John V was born on 4 September 1504 in
  Dessau and died on 4 February 1551 in Zerbst
- John V was the second but eldest surviving
  son of Ernest I, Prince of Anhalt-Dessau
- Mother was Margarete, daughter of Henry I,
  Duke of Munsterberg-Oels
- Great-grandson of George of Podebrady,
  King of Bohemia
- Married Margaret, daughter of Joachim I
  Nestor, Elector of Brandenburg
- Karl I, Prince of Anhalt-Zerbst was the
  eldest son of John V

## Related Pages
- [[people/Ernest-I-Prince-of-Anhalt-Dessau]]
  -- father of John V
- [[people/Karl-I-Prince-of-Anhalt-Zerbst]]
  -- eldest son and successor of John V

## Related Sources
- [[sources/digests/john-v-prince-anhalt-zerbst]]
  -- Wikipedia paragraph about John V
- [[sources/digests/karl-i-prince-of-anhalt-zerbst]]
  -- Wikipedia paragraph about Karl I
  mentioning John V
\end{verbatim}
\end{scriptsize}

\paragraph{Design Rationale.}
This schema supports query-time retrieval in three ways. First, aliases and tags allow \texttt{wiki\_search} to locate pages under different surface forms. Second, bidirectional wikilinks expose explicit traversal paths; for example, the question ``When did John~V's father die?'' can be answered by following the link from John~V to Ernest~I. Third, source references preserve provenance, allowing either the agent or a human auditor to verify claims against the original paragraph.

\section{Error Taxonomy}
\label{app:error_taxonomy}

Table~\ref{tab:error_types} lists the seven error categories used by the Error Book. Table~\ref{tab:error_distribution} further summarizes the distribution of these error types across compiled Wikis.

\begin{table}[h]
\centering
\small
\setlength{\tabcolsep}{4pt}
\renewcommand{\arraystretch}{1.08}
\begin{tabular}{p{2.7cm}p{4.3cm}}
\toprule
\textbf{Error Type} & \textbf{Description \& Detection} \\
\midrule
\multicolumn{2}{l}{\textbf{Structural validity errors}} \\
\midrule
Dangling Links & Inter-page links target non-existent pages; cross-validated with filesystem \\
\addlinespace[2pt]
Incomplete Pages & Required sections are missing (facts, sources); template completeness check \\
\addlinespace[2pt]
Malformed Refs & Source citations violate the format specification; regex validation \\
\addlinespace[2pt]
Unseen Overwrite & LLM modifies pages not selected in Step~1; set comparison \\
\addlinespace[2pt]
Index Inconsistency & Index--filesystem mismatch; bidirectional diff \\
\midrule
\multicolumn{2}{l}{\textbf{Content-level consistency errors}} \\
\midrule
Unsupported Facts & Page contains claims not grounded in the cited source digest; detected by source-grounded LLM verification \\
\addlinespace[2pt]
Cross-Page Contradictions & Related pages contain conflicting entity attributes, dates, or relations; detected by sampling-based consistency checks \\
\bottomrule
\end{tabular}
\caption{Error taxonomy spanning structural and content-level failures.}
\label{tab:error_types}
\end{table}

\paragraph{Cross-Corpus Error Distribution.}
Table~\ref{tab:error_distribution} reports the distribution of detected Error Book entries across compiled Wikis. Each cell denotes the percentage of detected errors of a given type within the corresponding Wiki, so each column sums to approximately 100\%.

\begin{table}[t]
\centering
\small
\setlength{\tabcolsep}{3.5pt}
\renewcommand{\arraystretch}{1.08}
\resizebox{\columnwidth}{!}{%
\begin{tabular}{lcccc}
\toprule
\textbf{Error Type} & \textbf{Hotpot} & \textbf{MuSiQue} & \textbf{2Wiki} & \textbf{AuthTrace} \\
\midrule
Dangling Links & 63.8\% & 62.5\% & 55.8\% & 29.1\% \\
Incomplete Pages & 2.3\% & 2.3\% & 2.1\% & 9.1\% \\
Malformed Refs & 20.5\% & 18.9\% & 22.3\% & 28.5\% \\
Unseen Overwrite & 3.5\% & 3.6\% & 2.6\% & 2.4\% \\
Index Inconsistency & 2.2\% & 2.5\% & 8.2\% & 10.9\% \\
Unsupported Facts & 5.8\% & 7.2\% & 6.4\% & 12.7\% \\
Cross-Page Contrad. & 1.9\% & 3.1\% & 2.6\% & 7.3\%  \\
\bottomrule
\end{tabular}%
}
\caption{Distribution of detected Error Book entries across compiled Wikis. Values are percentages within each Wiki.}
\label{tab:error_distribution}
\end{table}

\section{Fine-Grained and Additional Analyses}
\label{app:additional_analysis}

\begin{figure*}[h]
\centering
\includegraphics[width=\textwidth]{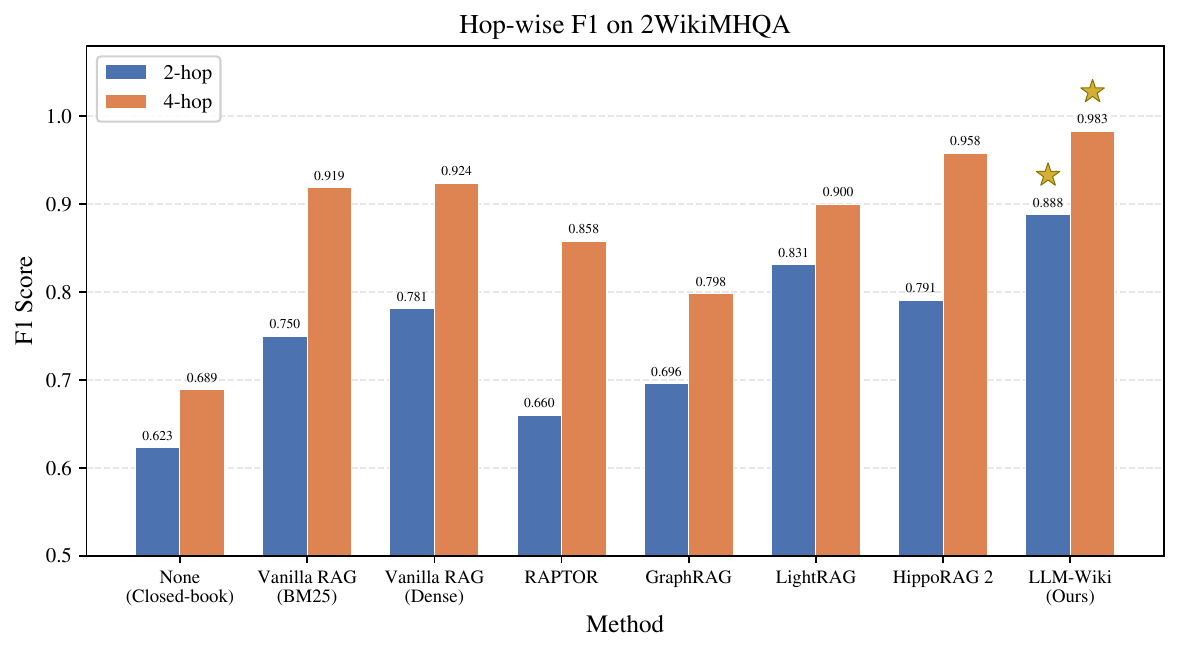}
\caption{Hop-wise F1 on 2WikiMHQA, grouped by reasoning depth (2-/4-hop).}
\label{fig:hopwise_app}
\end{figure*}

\paragraph{Hop-wise Breakdown.}
Figure~\ref{fig:hopwise_app} shows that LLM-Wiki's advantage increases with hop count. On 2WikiMHQA, the F1 gap between LLM-Wiki and the strongest baseline, LightRAG, widens from 5.7 F1 points on 2-hop questions to 8.3 F1 points on 4-hop questions. LLM-Wiki achieves 0.983 F1 on 4-hop questions, as pre-compiled bidirectional links reduce complex reasoning to compositional traversal, whereas Dense RAG reaches only 0.924 F1 on 4-hop questions due to the difficulty of capturing all intermediate entities through a single embedding query.

\paragraph{Type-wise Breakdown on 2WikiMHQA.}
Table~\ref{tab:typewise_app} reports performance across four 2WikiMHQA question types: bridge-comparison (Br.-Comp.), comparison (Comp.), compositional (Compos.), and inference (Infer.). Comparison questions approach ceiling performance at 0.989 F1 because pre-compiled entity pages place comparative attributes in an immediately accessible format. Compositional questions show the largest improvement, with LLM-Wiki outperforming Dense RAG by 15.6 F1 points and LightRAG by 7.3 F1 points, although the absolute score of 0.833 indicates room for further improvement on complex compositional reasoning.

\begin{table}[h]
\centering
\resizebox{\columnwidth}{!}{%
\begin{tabular}{lcccc}
\toprule
\textbf{Method} & \textbf{Br.-Comp.} & \textbf{Comp.} & \textbf{Compos.} & \textbf{Infer.} \\
\midrule
None (Closed-book) & 0.689 & 0.815 & 0.523 & 0.643 \\
Vanilla RAG (BM25) & 0.919 & 0.952 & 0.632 & 0.815 \\
Vanilla RAG (Dense) & 0.924 & 0.981 & 0.677 & 0.804 \\
RAPTOR & 0.858 & 0.934 & 0.532 & 0.642 \\
GraphRAG & 0.798 & 0.860 & 0.659 & 0.558 \\
LightRAG & 0.900 & 0.960 & 0.760 & 0.857 \\
HippoRAG~2 & 0.958 & 0.981 & 0.680 & 0.856 \\
\textbf{LLM-Wiki} & \textbf{0.983} & \textbf{0.989} & \textbf{0.833} & \textbf{0.909} \\
\bottomrule
\end{tabular}%
}
\caption{Type-wise F1 on 2WikiMHQA across four question categories.}
\label{tab:typewise_app}
\end{table}

\paragraph{Generalization Across Answer LLMs.}
To verify that LLM-Wiki's advantages are not specific to a particular answer model, we evaluate GPT-4o as an alternative answer LLM while keeping the compiled Wiki fixed (Table~\ref{tab:generalization}).

\begin{table}[h]
\centering
\small
\begin{tabular}{llccc}
\toprule
\textbf{Method} & \textbf{LLM} & \textbf{Hotpot} & \textbf{MuSiQue} & \textbf{2Wiki} \\
\midrule
Dense RAG & GLM-5.1 & 0.764 & 0.611 & 0.815 \\
\textbf{LLM-Wiki} & GLM-5.1 & \textbf{0.839} & \textbf{0.739} & \textbf{0.911} \\
\midrule
Dense RAG & GPT-4o & 0.741 & 0.503 & 0.636 \\
\textbf{LLM-Wiki} & GPT-4o & \textbf{0.792} & \textbf{0.608} & \textbf{0.805} \\
\bottomrule
\end{tabular}
\caption{Generalization across answer LLMs. The Wiki knowledge base compiled with GLM-5.1 is fixed; only the query-time answer LLM is changed.}
\label{tab:generalization}
\end{table}

Under GPT-4o, LLM-Wiki maintains substantial advantages: +5.1 F1 points on HotpotQA, +10.5 F1 points on MuSiQue, and +16.9 F1 points on 2WikiMHQA. This indicates that the benefits stem from knowledge organization and agent-native retrieval design rather than model-specific reasoning capabilities; structured traversal remains beneficial across answer models.

\paragraph{Efficiency Analysis.}
Table~\ref{tab:efficiency} compares query-time efficiency across methods.

\begin{table}[h]
\centering
\small
\begin{tabular}{lccc}
\toprule
\textbf{Method} & \textbf{Hotpot} & \textbf{MuSiQue} & \textbf{2Wiki} \\
& (s/q) & (s/q) & (s/q) \\
\midrule
None (Closed-book) & 24.7 & 30.0 & 21.6 \\
BM25 RAG & 18.1 & 30.2 & 16.7 \\
Dense RAG & 16.3 & 26.9 & 15.6 \\
RAPTOR & 19.9 & 40.2 & 28.0 \\
GraphRAG & 14.0 & 13.9 & 10.8 \\
LightRAG & 41.4 & 51.3 & 39.7 \\
HippoRAG~2 & 33.5 & 38.2 & 32.4 \\
\textbf{LLM-Wiki} & 14.9 & 27.1 & 15.9 \\
\bottomrule
\end{tabular}
\caption{Query latency in seconds per question on three benchmarks.}
\label{tab:efficiency}
\end{table}

LLM-Wiki achieves query latencies of 14.9s on HotpotQA and 15.9s on 2WikiMHQA, comparable to or faster than BM25 and Dense RAG, while being substantially faster than LightRAG, HippoRAG~2, and RAPTOR, which require 39.7--51.3s, 32.4--38.2s, and 28.0--40.2s, respectively. On MuSiQue, latency rises to 27.1s due to the dataset's greater complexity---2--4 hops requiring an average of 8.2 traversal steps---but remains well below LightRAG at 51.3s, RAPTOR at 40.2s, and HippoRAG~2 at 38.2s. GraphRAG achieves the lowest latency at 10.8--14.0s because it performs only a single community-summary lookup without iterative reasoning, but this efficiency comes at the cost of substantially lower accuracy. HippoRAG~2's higher latency, 32.4--38.2s, stems from its multi-stage pipeline: LLM-based query entity extraction, embedding-based entity linking, Personalized PageRank traversal, and a final LLM QA call with retrieved passages. The efficiency of LLM-Wiki stems from pre-compiled links that reduce runtime LLM calls: the agent reads an average of only 2.5--3.9 pages per query, compared with RAG's fixed 5 passages, while achieving higher accuracy through targeted traversal.

Counter-intuitively, the no-retrieval baseline is slower at 21.6--30.0s per question because the model must spend more generation time relying on parametric knowledge, whereas LLM-Wiki provides direct evidence that enables concise and confident generation.

\section{Detailed Case Studies and Retrieval Strategies}
\label{app:case_study}

We present detailed walkthroughs that illustrate the advantage of structured knowledge traversal over embedding-based retrieval.

\paragraph{Case 1: Bridge-comparison (4-hop)---Dense RAG fails, LLM-Wiki succeeds.}

We analyze a real 4-hop bridge-comparison question from 2WikiMultiHopQA: ``Which film has the director who is older, \textit{The Gamecock} or \textit{Monster A Go-Go}?'' Dense RAG retrieves the two film pages via embedding similarity, but fails to retrieve the director biography pages containing birth dates, which provide the evidence needed to answer the question. Its top-5 retrieved passages are \textit{The Gamecock (film)}, \textit{Monster a Go-Go}, \textit{The Mask of the Gorilla}, \textit{The Capture of Bigfoot}, and \textit{Monster from the Ocean Floor}, three of which are irrelevant. Forced to infer from film descriptions alone, it answers incorrectly: ``The Gamecock.''

The LLM-Wiki agent resolves the question in three steps: (1) it searches both film titles in parallel; (2) it batch-reads the two film pages and identifies director names from structured fields, namely Pasquale Festa Campanile for \textit{The Gamecock} and Bill Rebane and Herschell Gordon Lewis for \textit{Monster A Go-Go}; and (3) it follows the pre-compiled \texttt{links\_to} pointers to batch-read the director biography pages, obtaining exact birth dates: Herschell Gordon Lewis was born on 15 June 1926, whereas Pasquale Festa Campanile was born on 28 July 1927. Since Lewis (1926) is older than Campanile (1927), the correct answer is ``Monster A Go-Go.''

This case illustrates a core limitation of single-round embedding retrieval: it cannot reliably bridge the reasoning chain film $\to$ director $\to$ birth date because the intermediate entities are semantically distant from the original query. In contrast, LLM-Wiki makes this chain explicitly traversable through structured links.

\paragraph{Case 2: Link-following traversal---multi-hop reasoning via structural pointers.}
Question: ``When did John~V, Prince of Anhalt-Zerbst's father die?'' (from 2WikiMHQA)

Dense RAG must retrieve a passage that simultaneously mentions both John~V and his father's death date. In practice, the embedding query ``John V Prince of Anhalt-Zerbst father death'' retrieves passages about John~V's own biography or other Anhalt princes, but the father's death date resides in a separate passage about Ernest~I that may not mention John~V.

The LLM-Wiki agent resolves the question in three steps: (1) it searches ``John~V, Prince of Anhalt-Zerbst'' and locates the entity page; (2) it reads the page and finds a structured link to Ernest~I's page, annotated as ``father of John~V''; and (3) it follows this link and directly reads that ``Ernest~I died on 12 June 1516 in Dessau.'' The retrieval completes in three tool calls with no ambiguity. This demonstrates how pre-compiled bidirectional links transform multi-hop reasoning into compositional traversal: each hop is guided by explicit structural pointers rather than probabilistic similarity matching.

\paragraph{Retrieval Strategy Taxonomy.}
Beyond these specific cases, we observe three distinct retrieval strategies that the agent adaptively selects:

\begin{itemize}
    \item \textbf{Direct Retrieval}: For single-entity factual queries, the agent issues one \texttt{wiki\_search}, reads the target page, and extracts the answer directly, often completing the task in one or two tool calls.
    \item \textbf{Link-Following Traversal}: For multi-hop bridge queries, the agent reads entity pages and follows explicit \texttt{links\_to} pointers. Each hop is guided by structured links rather than probabilistic retrieval alone.
    \item \textbf{Browse \& Aggregation}: For open-ended or comparative queries, the agent reads directory index pages (\texttt{\_index.md}) to obtain a structured overview and then selectively batch-reads relevant pages. This strategy mirrors how humans browse a table of contents before focusing on relevant pages.
\end{itemize}

\end{document}